\documentclass[runningheads]{llncs}

\usepackage{eccv}

\usepackage{eccvabbrv}

\usepackage{graphicx}
\usepackage{booktabs}
\usepackage{mathbbol}
\usepackage{multirow}
\usepackage{algorithm}
\usepackage{algorithmic}
\usepackage[accsupp]{axessibility}  

\usepackage{hyperref}

\usepackage{orcidlink}

\begin{document}

\title{Generalizable Semantic Vision Query Generation for Zero-shot Panoptic and Semantic Segmentation} 

\titlerunning{Generalizable Semantic Vision Query Generation}

\author{Jialei Chen\inst{1} \and
Daisuke Deguchi\inst{1} \and
Chenkai Zhang\inst{1} \and
Hiroshi Murase\inst{1}}

\authorrunning{Jialei Chen et al.}

\institute{Nagoya University, Furo-cho, Chikusa-ku 464-8601, Japan }

\maketitle

\begin{abstract}
  Zero-shot Panoptic Segmentation (ZPS) aims to recognize foreground instances and background stuff without images containing unseen categories in training. Due to the visual data sparsity and the difficulty of generalizing from seen to unseen categories, this task remains challenging. To better generalize to unseen classes, we propose  \textbf{C}onditional t\textbf{O}ken alig\textbf{N}ment and \textbf{C}ycle tr\textbf{A}nsi\textbf{T}ion (\textbf{CONCAT}), to produce generalizable semantic vision queries. First, a feature extractor is trained by CON to link the vision and semantics for providing target queries. Formally, CON is proposed to align the semantic queries with the CLIP visual CLS token extracted from complete and masked images. To address the lack of unseen categories, a generator is required. However, one of the gaps in synthesizing pseudo vision queries, \ie, vision queries for unseen categories, is describing fine-grained visual details through semantic embeddings. Therefore, we approach CAT to train the generator in semantic-vision and vision-semantic manners. In semantic-vision, visual query contrast is proposed to model the high granularity of vision by pulling the pseudo vision queries with the corresponding targets containing segments while pushing those without segments away. To ensure the generated queries retain semantic information, in vision-semantic, the pseudo vision queries are mapped back to semantic and supervised by real semantic embeddings. Experiments on ZPS achieve a 5.2\% hPQ increase surpassing SOTA. We also examine inductive ZPS and open-vocabulary semantic segmentation and obtain comparative results while being 2 times faster in testing. 
\end{abstract}
\section{Introduction}
With the continuous advancement of deep learning technologies \cite{resnet,vggnet,transformer}, we have witnessed the rapid evolution of image segmentation. Image segmentation aims to assign each pixel or instance to the right category and can be grouped into semantic \cite{FCN,deeplabev3}, instance \cite{maskrcnn,solov2} and panoptic segmentation \cite{panopticsegmentation,panopticsegformer}. Though remarkable, to train a model for segmentation, huge amounts of high-quality data are required to collect, which is time-consuming and expensive, \eg, each image in the Cityscapes \cite{cityscapes} dataset takes 1.5 hours to make the annotations accounting for quality control. To tackle this challenge, Zero-shot classification \cite{zeroshotlearning1,zeroshotlearning2} is proposed to classify the categories that have never been trained. Inspired by these researches, zero-shot semantic segmentation \cite{zegformer,zegclip,clipteacher,spnet,cagnet,zeroshotsemantic} has achieved much success. However, due to the data sparsity and the difficulties in generalizing from seen to unseen categories, the progress of Zero-shot Panoptic Segmentation (ZPS) has been relatively modest thus far.

\begin{figure}[tb]
\centering
\begin{minipage}[h]{0.5\linewidth}
\centering
\includegraphics[width=1\linewidth]{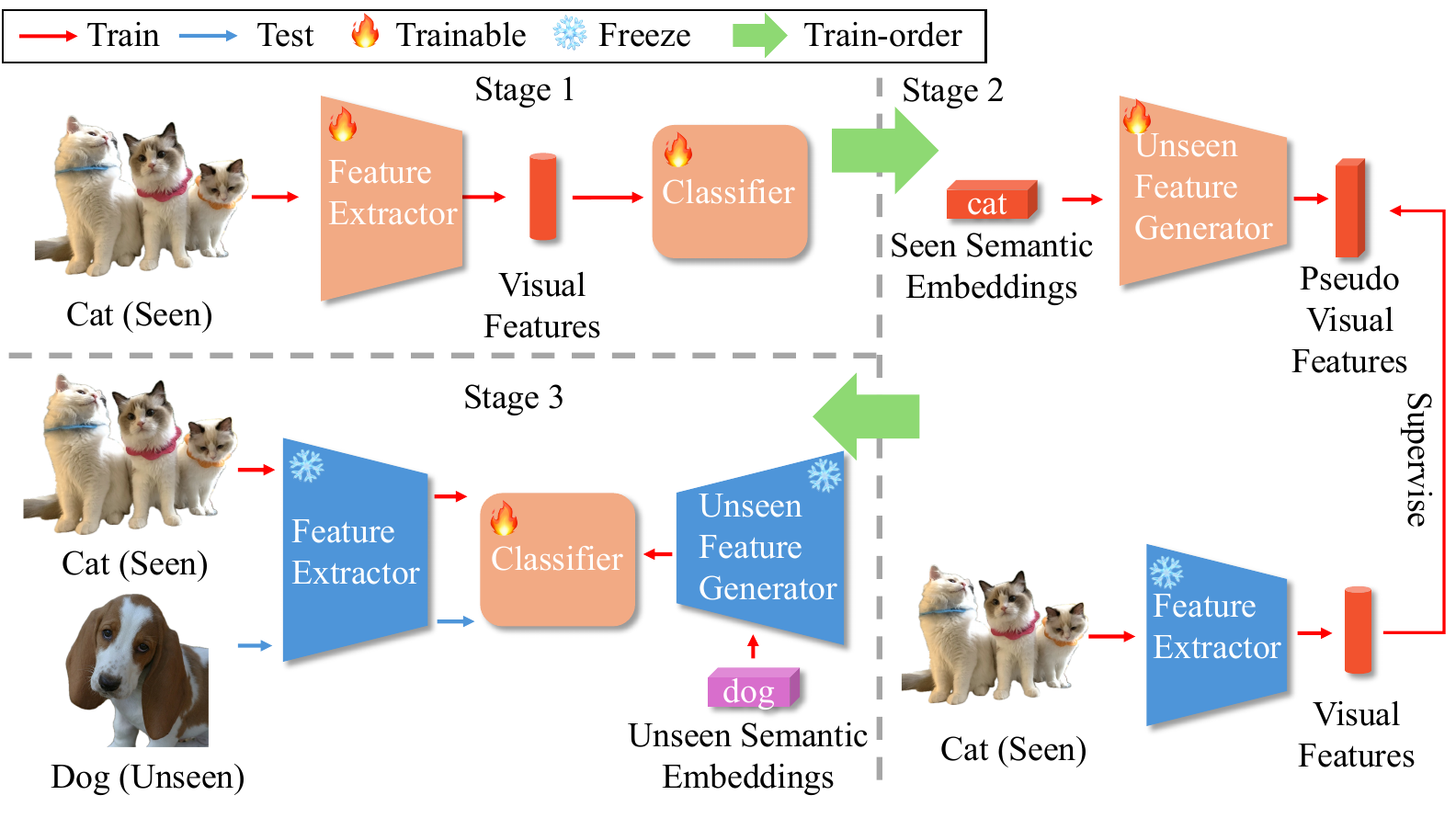}
\subcaption{Generation-based methods.}
\label{generation figure}
\end{minipage}
\hfill
\begin{minipage}[h]{0.49\linewidth}
\centering
\includegraphics[width=1\linewidth]{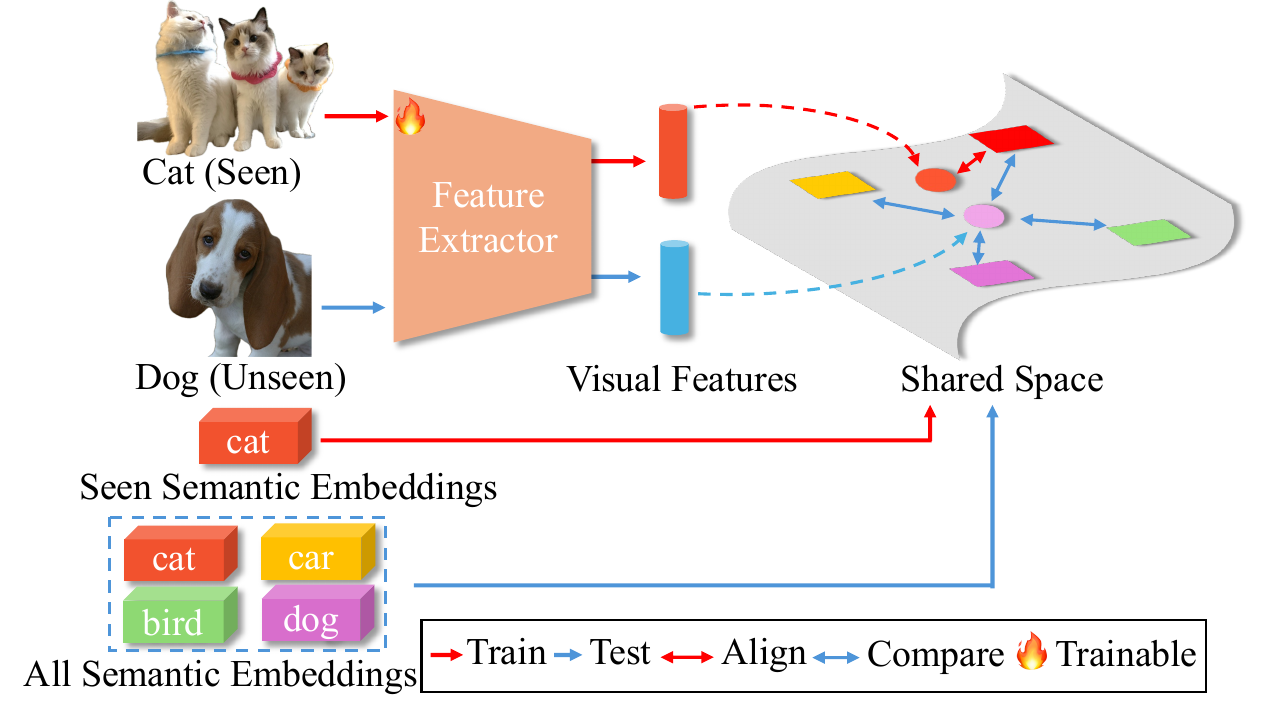}
\subcaption{Projection-based methods.}
\label{projection figure}
\end{minipage}
\vspace{-5pt}
\caption{Overview of the existing methods.}
\vspace{-20pt}
\end{figure}

Therefore, researchers apply generation-based methods \cite{zeroshotsemantic,cagnet,consistent} and projection-based \cite{zegclip,zegformer,spnet} to address the issues. For generation methods as shown in Figure \ref{generation figure}, the training stage is split into feature extractor training, unseen feature generator training, and classifier finetuning. First, a feature extractor with learnable classifiers is trained to fit the seen categories. Second, a generator is supervised under the trained feature extractor to synthesize unseen features from semantic embeddings. Finally, based on the generator and the feature extractor, the classifier can be finetuned by real seen queries and pseudo unseen queries. Projection-based methods, as shown in Figure \ref{projection figure}, map the model output and the seen semantic embeddings to a shared space, \eg, visual, semantic, during training, and recognize the unseen categories based on the generalization from seen to unseen, which establishes a cohesive link between semantic and visual.

However, both methods suffer from their shortages: \textbf{1)} for generation-based methods, the gap between the semantic and the vision impedes the quality of the pseudo features. \textbf{2)} For the projection-based methods, the model easily overfits the seen categories as the insufficiency of the information from unseen categories. Though some methods try to bridge the gap, \eg, PADing \cite{pading}, the link between vision and language is still weak, and the domain gap further impedes the quality of generated pseudo vision features. To address these issues, we propose \textbf{C}onditional t\textbf{O}ken alig\textbf{N}ment and \textbf{C}ycle tr\textbf{A}nsi\textbf{T}ion, \ie, \textbf{CONCAT} to produce generalizable semantic vision queries. CONCAT comprises two stages: training a feature extractor with a semantic projector by CON and training a generator that produce pseudo vision queries for unseen categories with CAT. 

Specifically, first, we replace \textbf{the learnable classifier with the semantic projector} to map the vision queries from the feature extractor to the semantic space represented by semantic embeddings, and the classification is calculated by the inner product between the semantic embeddings and semantic queries. Except for aligning vision and semantic embeddings by applying the semantic embeddings as the classifier \cite{zegclip,zegformer,spnet}, CON further enhances the relationship through the CLIP vision encoder. Formally, CON consists of two individual parts: Conditional Global Alignment (CGA) and Conditional Instance Alignment (CIA). For CGA, the predicted semantic queries are aligned with the CLIP visual CLS token extracted from the complete images. For CIA, the predicted semantic queries containing segments are aligned with the corresponding visual CLIP CLS token produced by the masked images. To provide information for unseen categories, previous researchers, however, directly map the semantic embeddings to the visual space while ignoring the domain gap between the semantics and vision \cite{pading,cagnet,sign,zeroshotsemantic,consistent}. Though some work \cite{pading} tried to reduce disparities, simply disentangling the pseudo vision queries into meaningful parts and noise can not reflect the high granularity of vision data. Consequently, we propose \textbf{C}ycle tr\textbf{A}nsi\textbf{T}ion (CAT) to train the generator in a semantic-vision and vision-semantic manner. Specifically, in semantic-vision, we map the semantic embeddings to the vision queries and learn the distribution of their corresponding real vision queries from the frozen feature extractor by Generative Moment Matching Network (GMMN) \cite{gmmn}. Moreover, motivated by the success of contrastive learning \cite{moco,simclr}, we propose Query Contrast (QC) to model the high granularity in vision by pulling the generated vision queries and the corresponding real ones together and pushing others \textbf{without segments} away. In vision-semantic, we project the generated vision queries back to semantics and supervise them with real semantic embeddings. After CONCAT, we union-finetune the semantic projector to adjust the unseen categories by both real and pseudo queries.

Our contributions can be listed as: 1) We propose CON which enhances the link of vision and semantics, serving as a base for the next two stages and the inductive ZPS. 2) We propose CAT which can generate generalizable semantic vision queries for unseen categories to conduct union-finetuning to achieve transductive ZPS. 3) Experiments on ZPS achieve a 5.2\% hPQ increase surpassing SOTA. We also examine inductive ZPS where only data related to seen categories can be accessed and open-vocabulary semantic segmentation and obtain comparative results while being \textbf{2 times faster} in testing. 

\section{Related Works}
\noindent \textbf{Panoptic Segmentation.} Panoptic segmentation \cite{panopticsegmentation} aims to assign the corresponding categories to each instance belonging to things and each pixel belonging to stuff simultaneously. Before the wide application of transformer \cite{transformer,vit} in computer vision, many works \cite{panopticfcn,upsnet} treat panoptic segmentation as the combination of instance segmentation \cite{maskrcnn,condinst} and semantic segmentation \cite{FCN,deeplabev3}. Since the proposal of DETR \cite{detr}, panoptic segmentation entered a new era. In a specific, first, the model is replaced by self-attention \cite{transformer,vit}. Moreover, the things and stuff are modeled together based on the Hungarian algorithm \cite{knet,panopticsegformer,maskformer,mask2former,detr,maskdino}. 

However, these methods still endure some drawbacks including the limitation of recognizable categories and the requirements for large amounts of data. To solve the first challenge, recently some works focusing on open-vocabulary panoptic segmentation have been proposed \cite{masqclip,openvocabularymaskclip,diffusionseg,freeseg,embeddingmodulation} where the model is aligned with the large-scale vision-language models, \eg, CLIP \cite{clip} and can generalize to the category that has never been seen during training. Unfortunately, these methods still require huge amounts of data with high-quality annotations.

\noindent \textbf{Zero-shot and Open-Vocabulary Segmentation.}
Another challenge that traditional image segmentation faces is the requirement of huge amounts of data which is expensive and time-consuming to collect. As a result, researchers devise two approaches to tackle segmentation tasks with limited data: zero-shot \cite{zegformer,zegclip,pading,sign,joint,strict,spnet,cagnet,simplebaseline,clipteacher} and open-vocabulary \cite{DeOP,simplebaseline,Openseg,Lseg,SAN,diffusionseg} segmentation is proposed where the model can recognize the category that has never been trained. Though these two approaches focus on transferring knowledge from seen to unseen categories, there are still some differences. First, open-vocabulary methods, \eg, SAN \cite{SAN}, train a feature extractor on a \textbf{complete} dataset and test on another one. However, zero-shot segmentation utilizes the images containing \textbf{only seen categories}, indicating much less data than open-vocabulary tasks. Second, Open-Vocabulary segmentation focuses on evaluating the generalization ability from a set of categories to another set where some similar or even the same categories may appear, \eg, car in both ADE20k \cite{ade20k} and COCO \cite{lin2014microsoft}, wall-concrete in COCO and wall in ADE20k. However, for zero-shot segmentation, the evaluation is within one dataset, and no overlap exists between the seen and unseen categories, testing if the unseen object can be recognized.

In this paper, the proposed CONCAT focuses on addressing the zero-shot panoptic segmentation. Compared with the most relevant work PADing \cite{pading}, we \textbf{replace the learnable classifier with the semantic embedding} and propose CON to provide a space with rich semantics. Additionally, rather than disentangling the pseudo vision queries into semantic-related and unrelated parts, we directly model the high granularity in vision space through vision query contrast. Compared with other tasks utilizing the images containing unseen segments, ZPS utilizes the images \textbf{without any unseen category}, so the pseudo labels can not be generated \cite{pseudoboundingboxlabels,crossmodallabeling,clipteacher}. We apply InfoNCE \cite{cpc} to transfer knowledge from CLIP rather than l1 loss \cite{hierarchicalvisual-languageknowledgedistillation,vild,globalknowledgecalibration} and CLS instead of dense tokens \cite{distillingdetr}.

\section{Methods}
\subsection{Preliminary}

\noindent \textbf{Task Formulation.} Before introducing our method, we need first to define the Zero-shot Panoptic Segmentation (ZPS). Given a dataset $\mathcal{D} = \left\{\textbf{X}^i,\textbf{Y}^i\right\}^M_{i=1}$ and the semantic embeddings $\textbf{A} \in \mathbb{R}^{N\times C}$ with channel $C$ of $N$ category in $\mathcal{D}$ where $\textbf{X} \in \mathbb{R}^{H\times W \times 3}$ denote the images with height $H$ and width $W$. $\textbf{Y}$ indicates the corresponding image label for $\textbf{X}$, $M$ is the number of images in a dataset. For ZPS, $\textbf{A}$ is split into two parts: seen categories $\textbf{A}_s$ and unseen categories $\textbf{A}_u$ where $\textbf{A}_s \cap \textbf{A}_u = \varnothing$. Besides, different from the Generalized Zero-Shot Semantic Segmentation (GZSSS) \cite{zegformer,zegclip,clipteacher} that maintains the images may contain the segments belonging to $\textbf{A}_u$, ZPS is trained by the images \textbf{with only the segments belonging to} $\textbf{A}_s$. This different setting leads to a sparse data scenario where a small fraction of data is retained, making ZPS more challenging \cite{pading}. Besides, inspired by the zero-shot semantic segmentation \cite{zegclip,clipteacher}, we split the ZPS into inductive ZPS where neither the $\textbf{A}_u$ nor images containing $\textbf{A}_u$ can be accessed, and transductive ZPS where only $\textbf{A}_u$ can be accessed in training.

\begin{figure}[t]
\begin{minipage}[h]{0.32\linewidth}
\centering
    \includegraphics[width=1\linewidth]{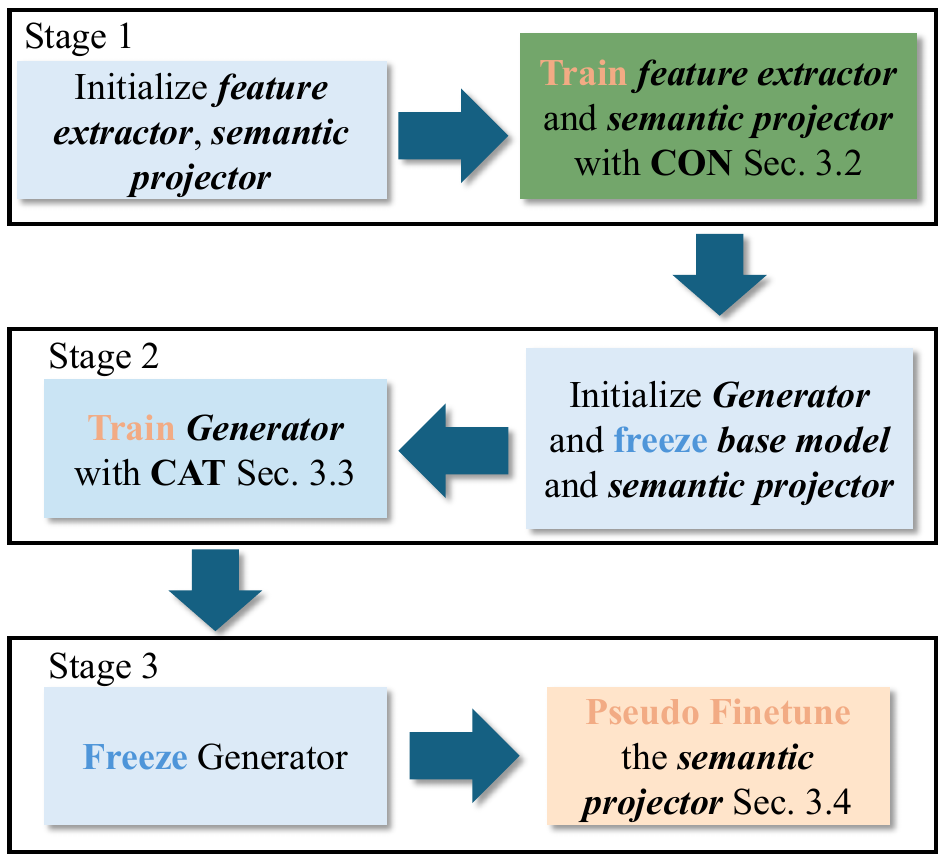}
    \vspace{-10pt}
    \caption{Training process.}
    \label{overview figure}
    \vspace{-10pt}
\end{minipage}
\hfill
\begin{minipage}[h]{0.66\linewidth}
    \centering
    \includegraphics[width=1\linewidth]{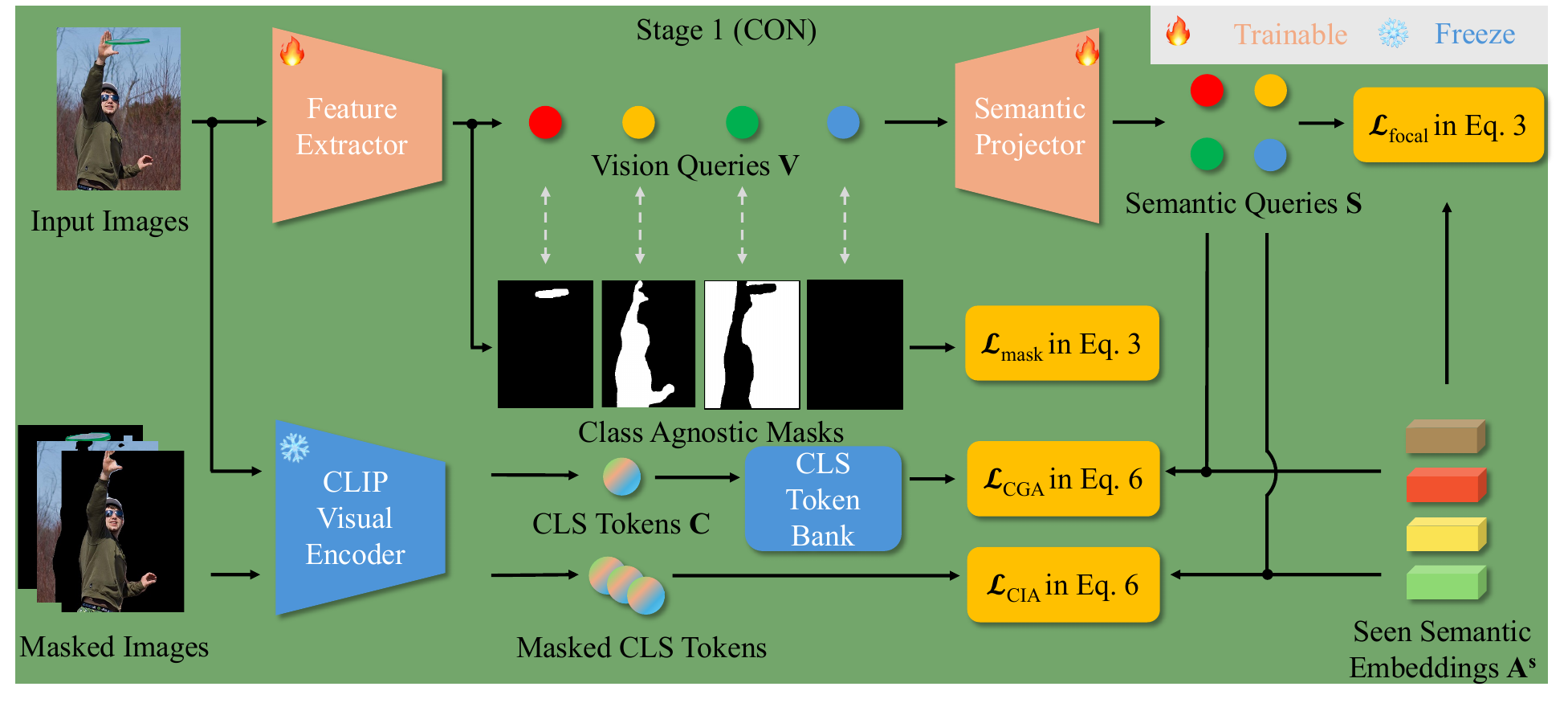}
    \vspace{-10pt}
    \caption{CON overview.}
    \label{CON figure}
    \vspace{-10pt}
\end{minipage}
\vspace{-15pt}
\end{figure}

\noindent \textbf{Method Overview.}
As CONCAT is based on the DETR-like \cite{detr} segmentor, the output of the feature extractor is called `queries' in this paper. Due to the data absence of unseen categories, CONCAT aims to produce generalizable pseudo semantic unseen queries. These data with the queries from images are utilized by union-finetuning to adjust segmentors to generalize from seen to unseen categories. Besides, we follow the paradigms of the generation-based method including training a feature extractor with a semantic projector \textbf{rather than a learnable classifier without any semantic}, training an unseen vision query generator, and union-finetuning the semantic projector as shown in Figure \ref{overview figure}. First, our model is separated into two parts: a feature extractor to extract visual queries and a semantic projector to map the vision queries to the semantic space to be semantic queries. Besides, aligning with the semantic embeddings, \ie, applying semantic embeddings as the classifier. Inspired by the work that transfers the knowledge from vision-language models \cite{globalknowledgecalibration,exploringfromclipvisionencoder,distillingdetr,clipteacher}, we propose the Conditional tOken aligNment (CON) for providing a shared space with rich semantics. CON consists of two parts: Conditional Global Alignment (CGA) and Conditional Instance Alignment (CIA). In CGA, the output semantic queries are aligned with the CLS token extracted by the CLIP visual encoder from a complete image, and the queries with segments are assigned with larger weights. In CIA, the queries containing a segment are aligned with the CLIP visual CLS tokens of the corresponding masked images. Second, though many works \cite{pading,cagnet,consistent} devoted efforts to generate high-quality pseudo vision queries with semantic embeddings, they fail to consider the high granularity of vision queries. Consequently, we propose Cycle trAnsiTion (CAT) where a generator is trained in a semantic-vision and vision-semantic manner. In semantic-vision, we propose query contrast to model the high granularity of vision queries through contrastive learning. In vision-semantic, the generated vision queries are mapped back to semantic and supervised by real semantic embeddings. Finally, only the semantic projector is union-finetuned by the pseudo vision queries form the frozen generator and real vision queries feature extractor.

\subsection{Conditional Token Alignment (CON)}\label{CON}
CLIP \cite{clip} has impressively linked vision and language, and shown remarkable ability in the zero-shot tasks. As a result, many works try to transfer the knowledge of CLIP through the CLS token from CLIP visual encoder \cite{PACL,clipteacher,maskawareclip,exploringfromclipvisionencoder}. Inspired by these works, we propose the Conditional tOken aligNment (CON) to align the semantic queries and the CLIP visual CLS tokens under the condition if a semantic query is assigned segments as shown in Fig. \ref{CON figure}. Therefore, CON offers bases in providing semantic vision queries for the next two stages. CON consists of Conditional Global Alignment (CGA) and Conditional Instance Alignment (CIA), which match the CLIP visual CLS tokens in different ways. CGA conditions on the combination of semantic queries to reconstruct CLS tokens and CIA conditions on whether a semantic query is assigned with segments.

Specifically, given $\textbf{X}$ and its corresponding label $\textbf{Y}$ containing segments, we feed $\textbf{X}$ to the feature extractors and get the vision queries $\textbf{V} \in \mathbb{R}^{K*D}$. Following the steps of mask-level classification methods, \eg, MaskFormer \cite{maskformer}, each $\textbf{v} \in \textbf{V}$ predicts its corresponding masks $\textbf{m} \in \textbf{M} \in [0,1]^{K\times H \times W}$ where $K$ indicates the number of vision queries and $D$ is the dimension of vision queries. Then, to link the vision and semantic, $\textbf{V}$ is fed into the semantic projector to obtain the semantic queries $\textbf{S} \in \mathbb{R}^{K \times C}$. The classification logits of masks $\textbf{M}$ are:
\begin{equation}
    P(\textbf{S}, \textbf{A}_s) = sigmoid(\textbf{S} \textbf{A}^T_s),
    \label{get probability}
\end{equation}
where $P \in [0,1]^{K \times N}$ indicates the intensity of the response of each mask to the seen categories. Motivated by the success of bipartite matching-based assignment \cite{detr,mask2former,maskformer}, a matching $\hat{\sigma}$ that assign each $\textbf{s}$ to the ground truth is calculated by:
\begin{equation}
    \hat{\sigma} = \arg\min_{\sigma}\sum_{s_i \in \textbf{S}}\mathcal{L}_{match}(y_{\sigma(s_i)}, s_i),
    \label{matching}
\end{equation}
\begin{equation}
    \mathcal{L}_{match}(y_{\sigma(s_i)},s_i) = \mathcal{L}_{focal}(p_{\sigma(s_i)},p_{s_i}) +  \mathcal{L}_{mask}(m_{\sigma(s_i)},m_{s_i}),
    \label{normal loss}
\end{equation}
where $\mathcal{L}_{focal}$ implies the focal loss proposed in \cite{focalloss,chen2022uncertainty,clipteacher}, $\mathcal{L}_{mask}$ is the mask loss applied in \cite{mask2former,maskformer}. $p_{s_i}$ and $m_{s_i}$ are the prediction of category and mask from the $i$th semantic and vision query. $y_{\sigma(s_i)}$ indicates the labels (category $p_{\sigma(s_i)}$ and mask $m_{\sigma(s_i)}$) assigned to $s_i$ under the assignment $\sigma$. $p_{\sigma(s_i)}$ and $m_{\sigma(s_i)}$ are the label for $s_i$ under an assignment $\sigma$. Note that we replace the widely used cross entropy \cite{zegformer,mask2former,maskformer} with focal loss \cite{focalloss,chen2022uncertainty,clipteacher} to avoid overfitting seen categories.

However, with only $\mathcal{L}_{match}$ the output of the segment model is only aligned with the semantic embeddings leading to sub-optimal performance. To further align with the CLIP model, we propose CON composed of CGA and CIA. Specifically, we first freeze the CLIP visual encoder to avoid the loss of the strong alignment between vision and language. Besides feeding into the feature extractor and semantic projector, $\textbf{X}$ also needs to be put into the frozen CLIP encoder to obtain the visual CLS token $\textbf{C} \in \mathbb{R}^{1\times C}$. Then, we compute the the similarity $\textbf{W} \in [0,1]^{K \times 1}$ between semantic queries $\textbf{S}$ and CLS tokens $\textbf{C}$,
\begin{equation}
    \textbf{W} = softmax(\frac{\textbf{S}^T \textbf{C}}{||\textbf{S}||_2 \cdot ||\textbf{C}||_2} \cdot (\gamma \cdot \mathbb{1}{\left\{y_{\hat{\sigma}(s_i)} \neq \varnothing\right\}} + 1)), s_i \in S,
    \label{CLS similarity}
\end{equation}
where $||\cdot||_2$ indicates the $\ell_2$ norm of a vector. $\gamma$ is a hyperparameter to control the scale of the similarity. Next, based on $\textbf{W}$, the conditional reconstructed CLS token $\hat{\textbf{C}} \in \mathbb{R}^{1\times C}$ is obtained by
\begin{equation}
    \hat{\textbf{C}} = \textbf{W}^T \textbf{S}.
    \label{reconstructed CLS}
\end{equation}
Motivated by the success of contrastive learning \cite{moco,simclr,mocov2} and InfoNCE \cite{cpc},  CGA is introduced to further bridge the gap between feature and semantics,
\begin{equation}
  \mathcal{L}_{CGA}(\textbf{C},\hat{\textbf{C}}) = \Sigma_i^B{\frac{\exp(\textbf{c}^T_i \hat{\textbf{c}}_i / \tau)}{\Sigma_{j \neq i}^{B}{\exp(\textbf{c}_{j}^T \hat{\textbf{c}}_i) / \tau})+\exp(\textbf{c}^T_i \hat{\textbf{c}}_i / \tau)}},
  \label{CGA formula}
\end{equation}
where $c_i \in \textbf{C}$ and $\hat{c_i} \in \hat{\textbf{C}}$. $B$ indicates the batch size of the image, and $\tau$ is the temperature parameter. To further align the feature extractor and the CLIP vision model, we also apply the token bank \cite{clipteacher} $\textbf{B}$ to replace the $\textbf{C}$ in $\mathcal{L}_{CGA}$ where the feature extractor can align with the CLIP through more CLS tokens.

Inspired by the recent success of transferring the knowledge of CLIP through the CLS tokens \cite{distillingdetr,exploringfromclipvisionencoder,PACL}, we propose CIA to transfer the CLIP knowledge better. Formally, given $\textbf{X}$ and its corresponding panoptic label $\textbf{Y}$, we first get all the segments $\textbf{y}_i$ in $\textbf{Y}$ where $\textbf{Y} = \sum_{i}^O\textbf{y}_i$ and $O$ indicates the number of segments in $\textbf{Y}$. Then, we achieve the masked image $\textbf{X}'$ through the multiplication between the $\textbf{y}$ and $\textbf{X}$. Next, the masked images are put into the CLIP visual encoder to get the visual CLS token of masked images $\textbf{C}' \in \mathbb{R}^{O*C}$. The semantic queries $\textbf{S}'$ corresponding to segments are obtained through $\hat{\sigma}$.
\begin{equation}
    \textbf{S}' = \left\{s' \in \textbf{S}|y_{\hat{\sigma}(s')} \neq \varnothing \right\}.
    \label{object queries}
\end{equation}
Finally $\mathcal{L}_{CIA}$ is computed between the $\textbf{S}'$ and $\textbf{C}'$,
\begin{equation}
  \mathcal{L}_{CIA}(\textbf{C}',\textbf{S}') = \Sigma_i^O{\frac{\exp(\textbf{c}^{'T}_{i}\textbf{s}^{'}_i / \tau)}{\Sigma_{j \neq i}^{O}{\exp(\textbf{c}_{j}^{'T}  \textbf{s}_i^{'}) / \tau})+\exp(\textbf{c}^{'T}_i  \textbf{s}_i^{'} / \tau)}}.
  \label{CIA formula}
\end{equation}
The loss $\mathcal{L_{\text{1}}}$ with a hyperparameter $\lambda_c$ to control CIA for the first stage is,
\begin{equation}
    \mathcal{L_{\text{1}}} = \mathcal{L}_{CGA}(\textbf{C},\hat{\textbf{C}}) +  \lambda_c\mathcal{L}_{CIA}(\textbf{C}',\textbf{S}') + \mathcal{L}_{match}(y_{\hat{\sigma(\textbf{S})}},\textbf{S})).
    \label{first total loss}
\end{equation}

\begin{figure}[t]
\begin{minipage}[h]{0.585\linewidth}
    \centering
    \includegraphics[width=1\linewidth]{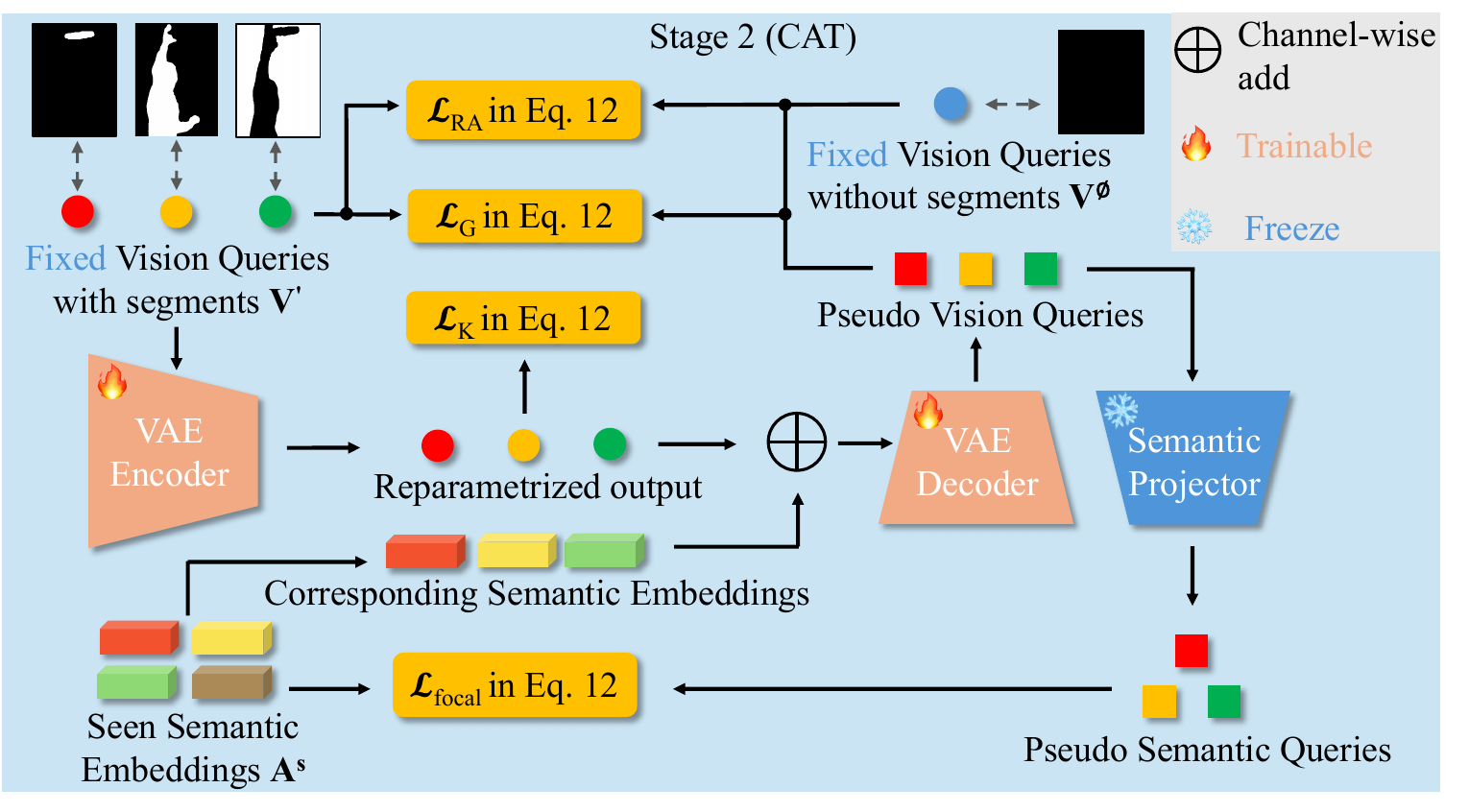}
    \vspace{-15pt}
    \caption{CAT overview where $y_{\hat{\sigma}(\textbf{V}^{\varnothing})} = \varnothing$.}
    \label{CAT figure}
    \vspace{-15pt}
\end{minipage}
\hfill
\begin{minipage}[h]{0.38\linewidth}
\centering
    \includegraphics[width=1\linewidth]{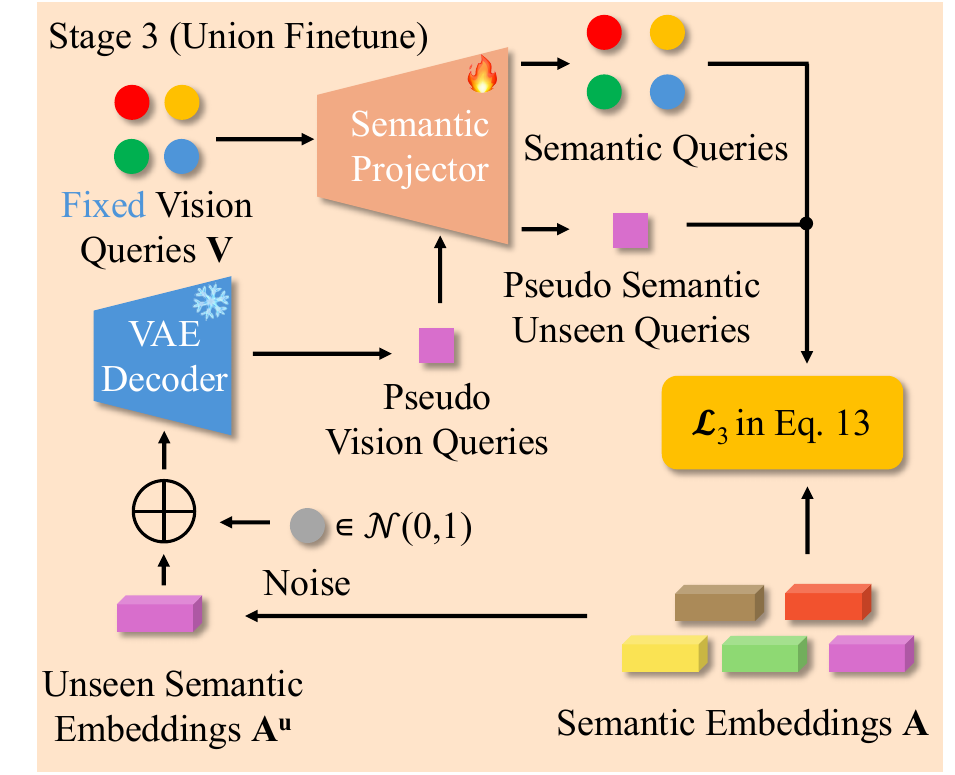}
    \vspace{-15pt}
    \caption{Stage 3 overview.}
    \label{finetune figure}
    \vspace{-15pt}
\end{minipage}
\end{figure}
\subsection{Cycle Transition (CAT)}\label{CAT}

As the sparsity of the unseen data, one of the greatest challenges for the projection-based \cite{zegformer,zegclip,clipteacher} methods is overfitting to the seen categories. Consequently, we follow the generation-based \cite{pading,cagnet} to train a generator to provide pseudo unseen from semantic embeddings to improve the performance of the model in the first stage \cite{cagnet,zeroshotsemantic}. However, as the gap between the semantic and visual space, \ie, one visual feature contains lots of semantic-unrelated clues, directly generating from semantic to visual may lead to unsatisfactorily feature quality. As a result, PADing \cite{pading} disentangles the generated features to semantic-related and noise parts. Unfortunately, we believe that the visual features are not fully utilized, especially those corresponding to no segments. So, we propose Cycle tr\textbf{A}nsi\textbf{T}ion (CAT) to further improve the quality of the pseudo queries as shown in Fig. \ref{CAT figure}.

CAT trains the generator in a semantic-vision and vision-semantic manner. Before CAT, we freeze all the parameters of the feature extractor and the semantic projector and apply the Conditional VAE \cite{bvae} as the generation model. In the semantic-vision, given $\textbf{X}$, we only obtain the fixed vision queries corresponding to segments $\textbf{V}'$ and the corresponding semantic embeddings $\textbf{A}_v \subset \textbf{A}_s$ based on the $\hat{\sigma}$. Then, we feed the $\textbf{V}'$ to the CVAE encoder to obtain the latent variables $\textbf{Z}'$ after reparameterization \cite{vae}. Next, we add the condition $\textbf{A}_v$ and the $\textbf{Z}'$ to get the conditional latent variables $\textbf{Z}$. Finally, $\textbf{Z}$ is put into the CVAE decoder and produces the pseudo seen visual queries $\hat{\textbf{V}}'$. Then, to mimic the distribution of the real seen visual queries, we follow the Generative Moment Matching Network (GMMN) \cite{gmmn} to reduce the maximum mean discrepancy between them,
\begin{equation}
    \mathcal{L}_{g} = \sum_{\hat{\textbf{v}}' \in \hat{\textbf{V}}'}k(\textbf{V}', \hat{\textbf{v}}') + \sum_{\textbf{v}' \in \textbf{V}'}k(\textbf{v}', \hat{\textbf{V}}') - 2\sum_{\hat{\textbf{v}}' \in \hat{\textbf{V}}'} \sum_{\textbf{v}' \in \textbf{V}'}k(\textbf{v}', \hat{\textbf{v}}'),
    \label{generative loss}
\end{equation}
where $k(\textbf{v}', \hat{\textbf{v}}') = \exp{(\frac{1}{2\lambda^2} ||\textbf{v}'-\hat{\textbf{v}}'||^2)}$ is a kernel function with a bandwidth $\lambda$.

To further improve the qualities of the produced pseudo vision queries, we propose query contrast. Specifically, given $\textbf{V}$, $\textbf{V}'$ and $\hat{\textbf{V}}'$ based on the best matching $\hat{\sigma}$, we find out the queries $\textbf{V}^{\varnothing}$ that are corresponding to $\varnothing$ where $\textbf{V}' \cup \textbf{V}^{\varnothing} = \textbf{V}$. Then inspired by the InfoNCE \cite{cpc}, we pull the $\hat{\textbf{V}}'$ and the corresponding $\textbf{V}'$ together and push $\textbf{V}_n$ apart,
\begin{equation}
  \mathcal{L}_{RA} = \Sigma_i^O{\frac{\exp(\textbf{v}^{'T}_i  \hat{\textbf{v}}'_{i} / \tau_r)}{\Sigma_{i}^{O}{\exp(\textbf{v}_{j}^{'T} \hat{\textbf{v}}'_i) / \tau_r})+\Sigma_{j}^{K - O}{\exp(\textbf{v}^{\varnothing T}_i  \hat{\textbf{v}}'_i / \tau_r)}}},
  \label{query contrast}
\end{equation}
where $\tau_r$ is the temperature parameter to control the scale of query contrast.

In the vision-semantic, we forward $\hat{\textbf{V}}'$ to the fixed semantic project to obtain $\hat{\textbf{S}}'$. Then the $\hat{\textbf{S}}'$ is aligned with the $\textbf{A}'$ through the classification loss in Eq. \ref{normal loss}, \ie, focal loss. In this way, the pseudo vision queries possess category attributes, facilitating subsequent finetuning. The loss in the second stage is:
\begin{equation}
    \mathcal{L_{\text{2}}} = \mathcal{L}_{g} + \mathcal{L}_{k} + \lambda_r \mathcal{L}_{RA} + \lambda_f \mathcal{L}_{focal}(\hat{\textbf{P}}',\textbf{P}_{\hat{\sigma}}).
    \label{second total loss}
\end{equation}
where $\mathcal{L}_{k}$ forces the latent embedding to obey the normal distribution \cite{vae,bvae}. $\lambda_r$ and $\lambda_f$ are two hyperparameters to control the scale of corresponding loss.

\subsection{Union-Finetuning} \label{finetune sec}
Different from most of the generation-based methods \cite{pading,sign,cagnet} that finetune the learnable classifier without semantics, we finetune the semantic projector while keeping the classifiers, \ie, semantic embeddings $\textbf{A}$, unchanged as shown in Fig. \ref{finetune figure}. Specifically, we fix the feature extractor and the generator trained in the first and second stages and make the semantic projector trainable. Given $\textbf{X}$, the semantic embeddings for the unseen categories $\textbf{A}_u$ and the semantic embeddings $\textbf{S}$, we first randomly sample some noise $\textbf{U} \in \mathcal{N}^{U*D}$ where $U$ is the number of noise features. Then randomly sample $U$ unseen semantic embeddings from $\textbf{A}_u$ and feed the combination with $\textbf{U}$ to the frozen CVAE decoder to obtain the pseudo unseen vision queries. Finally, these pseudo unseen vision queries are put into the semantic projector to get the pseudo unseen semantic unseen $\hat{\textbf{S}}'_u$. The semantic projector is union-finetuned by $\textbf{S}$ and $\hat{\textbf{S}}'_u$:
\begin{equation}
    \mathcal{L_{\text{3}}} = \mathcal{L}_{focal}(P(\textbf{S},\textbf{A}_s),y_{\hat{\sigma}(\textbf{S})}) + \mathcal{L}_{focal}(P(\hat{\textbf{S}}'_u,\textbf{A}), l_u).
    \label{finetune}
\end{equation}
where $l_u$ implies the index in all the semantic embeddings (including the unseen ones) for the random selection from $\textbf{A}_u$. $y_{\hat{\sigma}(\textbf{S})}$ are labels for semantic queries $\textbf{S}$.

\section{Experiments}

\subsection{Experiments Setup}
\noindent \textbf{Dataset.} The dataset we use is COCO 2017 \cite{lin2014microsoft} which contains 118K images for training, 5K images for validation, and 133 categories (80 things and 53 stuff). Following PADing \cite{pading}, we choose 73 things and 46 stuff as the seen categories, and the training procedure applies the images with only seen categories, therefore, only \textbf{45,617 (45K) images are kept less than 118k}.

\noindent \textbf{Implementation Details.} For panoptic tasks, we leverage MMDetection \cite{mmdetection}, and for semantic tasks, we utilize MMSegmentation \cite{mmsegmentation} on 8 V100 GPUs. semantic embeddings are extracted using the CLIP text encoder \cite{clip} with a ViT-B/16 \cite{vit} backbone. The text templates align with prior works \cite{zegformer, clipteacher, zegclip, pading}. Our feature extractor, Mask2Former \cite{mask2former}, employs a ResNet-50 \cite{resnet} backbone with the same hyperparameters as the original Mask2Former. The semantic projector is a simple MLP, and the CVAE structure consists of four layers of MLP-Batchnorm1d-leakyReLU, serving as both encoder and decoder. The generator optimizer aligns with the feature extractor but with a learning rate ten times higher. In union-finetuning, the semantic projector's learning rate is set at 0.1 times the feature extractor. The feature extractor undergoes default training for 48 epochs (140,350 iterations), considerably less than the original Mask2Former (368,750 iterations). In CAT, the model trains for 12 epochs (36,875 iterations). Hyperparameters include $\gamma = 2$, $\tau = 0.07$, $\lambda = \left\{2, 5, 10, 20, 40, 60\right\}$, $\lambda_r = 0.1$, $\lambda_f = 0.01$, bank size in CGA is 32. During inference, the segment threshold is 0.1, and in inductive settings, the logit for unseen categories is incremented by 1, and no other increments for transductive settings. For union-finetuning, we only report the best performance as different settings lead to different convergence speeds. Other hyperparameters can be seen in the \textit{\textbf{Supplementary materials}}. 

\noindent \textbf{Evaluation Metric.} We follow the GZSL settings \cite{pading,zegformer,zegclip,clipteacher} where both the seen and the unseen categories need to be segmented correctly. As a result, during inference, both $\textbf{A}_s$ and $\textbf{A}_u$ are utilized as the classifier. To comprehensively consider the performance for both seen and unseen categories, we apply the harmonic panoptic quality (hPQ) for panoptic segmentation and Intersections over Union (hIoU) for semantic segmentation as the evaluation metric,
\begin{minipage}{0.45\textwidth}
\begin{equation}
    hPQ = \frac{2 \cdot sPQ \cdot uPQ}{sPQ + uPQ},
\label{hpq}
\end{equation}
\end{minipage}
\vspace{5pt}
\hfill
\begin{minipage}{0.45\textwidth}
\begin{equation}
    hIoU = \frac{2 \cdot sIoU \cdot uIoU}{sIoU + uIoU},
\label{hiou}
\end{equation}
\end{minipage}
where $sPQ$ and $uPQ$ denote the panoptic quality \cite{panopticsegmentation} for the seen and unseen categories. $sIoU$ and $uIoU$ denote the mIoU for the seen and unseen categories.

\begin{table}[t]
\setlength{\tabcolsep}{20pt}
\caption{Comparison with other zero-shot panoptic methods. }
\vspace{-10pt}
\resizebox{\linewidth}{!}{
\begin{tabular}{c|ccc|ccc}
\toprule
\multirow{2}{*}{Model}    & \multicolumn{3}{c|}{Inductive}                                                                    & \multicolumn{3}{c}{Transductive}                                                                 \\ \cmidrule{2-7} 
                          & hPQ                               & sPQ                      & uPQ                                & hPQ                               & sPQ                      & uPQ                               \\ \midrule
PADing \cite{pading}                  & 0.0                               & \textbf{43.3}            & 0.0                                & 22.3                              & \textbf{41.5}            & 15.3                              \\
\multicolumn{1}{l|}{Ours} & \multicolumn{1}{l}{\textbf{20.8}} & \multicolumn{1}{l}{40.5} & \multicolumn{1}{l|}{\textbf{14.0}} & \multicolumn{1}{l}{\textbf{27.5}} & \multicolumn{1}{l}{39.2} & \multicolumn{1}{l}{\textbf{21.2}} \\ \bottomrule
\end{tabular}
}
\label{ZPS comparison}
\vspace{-5pt}
\end{table}

\begin{table}[t]
\caption{Comparison on zero-shot semantic segmentation task.}
\vspace{-10pt}
\setlength{\tabcolsep}{17pt}
\resizebox{\linewidth}{!}{
\begin{tabular}{ccc|cccc}
\toprule
Method    & Backbone                    & Embed    & sIoU          & uIoU          & hIoU  & FPS        \\ \midrule
SPNet \cite{spnet}    & \multirow{6}{*}{ResNet-101 \cite{resnet}} & Word2vec & 35.2          & 8.7           & 14.0   & -       \\
ZS3 \cite{zs3}      &                             & Word2vec & 34.7          & 9.5           & 15.0        & -  \\
CaGNet \cite{cagnet}   &                             & Word2vec & 33.5          & 12.2          & 18.2        & -  \\
SIGN \cite{sign}     &                             & Word2vec & 32.2          & 15.5          & 20.9       & -   \\
Zsseg-seg \cite{simplebaseline}&                             & CLIP     & 38.7          & 4.9           & 8.7  & 1.11          \\
ZegFormer \cite{zegformer} &                             & CLIP     & 37.4          & 21.4          & 27.2   & 6.69       \\ \cmidrule{1-3}
DeOP \cite{DeOP} &    \multirow{2}{*}{ResNet-101c \cite{deeplabev3}}                         & CLIP     & 38.0          & 38.4          & 38.2    & 4.37      \\
Ours &                          & CLIP     & 40.0          & \textbf{38.6} & \textbf{39.3} & 9.24     \\ \cmidrule{1-3}
PADing \cite{pading}   & \multirow{2}{*}{ResNet-50 \cite{resnet}}  & CLIP     & \textbf{40.4} & 24.8          & 30.7    & -      \\
Ours      &                             & CLIP     & 40.1          & 32.9 & 36.1 & \textbf{24.0} \\ \bottomrule
\end{tabular}
}
\label{comparison ZSS}
\vspace{-10pt}
\end{table}

\subsection{Comparison with other methods}
\noindent \textbf{Comparison on ZPS. }We commence by comparing our method with state-of-the-art Zero-Shot Panoptic Segmentation (ZPS) techniques in both inductive and transductive settings, presenting the results in Tab. \ref{ZPS comparison}. In the inductive scenario, where the primitive generator in PADing \cite{pading} and the second stage in our method are directly inferred on the test dataset without training, we observe that PADing achieves 0 hPQ, with its sPQ surpassing ours by 2.8\%, attributed to overfitting on the seven categories. However, our approach excels, surpassing PADing by \textbf{14\%} in uPQ, showcasing the merits of providing a shared space with rich semantics. In the transductive settings, our method outperforms PADing by 5.2\% hPQ and 2.0\% uPQ, even if the sPQ is slightly lower than PADing.

\noindent \textbf{Comparison on Zero-shot Semantic Segmentation (ZSS).} We conducted experiments on challenging ZSS tasks. The metric we use is hIoU in DeOP \cite{DeOP} and the dataset is COCO \cite{coco}. It's important to note that the reported performance does not involve self-training or model ensemble with CLIP visual encoder. As depicted in Tab. \ref{comparison ZSS}, in comparison to PADing, our sIoU performance is slightly lower (40.0\% vs. 40.4\%). However, our uIoU significantly surpasses PADing by 8.1\%. The hIoU, representing the overall performance of both sIoU and uIoU, is 5.4\% higher than PADing. It's worth mentioning that these results are achieved using the ResNet-50 \cite{resnet}, and our method even outperforms some approaches, such as ZegFormer \cite{zegformer} with ResNet-101. Compared with DeOP \cite{DeOP}, we can achieve higher performance while 2$\times$ faster. 

\begin{table}[tb]
\vspace{-5pt}
\setlength{\tabcolsep}{8pt}
\caption{Ablation study on the proposed methods. ``S'' indicates the softmax plus cross-entropy as the loss function and ``F'' indicates the focal loss. ``G'' indicates global token alignment ($\gamma = 0$) and ``CG'' indicates the proposed CGA. ``Sup'' indicates the supervision by the real semantic embeddings in the second stage, ``Fcls'' indicates if the first $\mathcal{L}_{focal}$ exists in the third stage. ``QC'' indicates the query contrast in the second stage. PADing indicates the procedure in \cite{pading}.}
\resizebox{\linewidth}{!}{
\begin{tabular}{cccccc|ccc|ccc}
\toprule
\multicolumn{6}{c|}{Proposed Method}                                               & \multicolumn{3}{c|}{Inductive}                                        & \multicolumn{3}{c}{Transductive}                                   \\ \midrule
\multicolumn{3}{c|}{CON}              & \multicolumn{2}{c|}{CAT}        & Finetune & \multirow{3}{*}{hPQ}  & \multirow{3}{*}{sPQ}  & \multirow{3}{*}{uPQ}  & \multirow{3}{*}{hPQ} & \multirow{3}{*}{sPQ} & \multirow{3}{*}{uPQ} \\ \cmidrule{1-6}
S/F & CG/G & \multicolumn{1}{c|}{CIA} & QC  & \multicolumn{1}{c|}{Sup} & Fcls     &                       &                       &                       &                      &                      &                      \\ \cmidrule{1-6}
\multicolumn{6}{c|}{First Stage}                                                   &                       &                       &                       &                      &                      &                      \\ \midrule
F   & -    & -                        & $\checkmark$ & $\checkmark$                       & $\checkmark$      & 10.9                  & 31.0                  & 6.6                   & 18.1                 & 28.7                 & 13.2                 \\
F   & G    & -                        & $\checkmark$ & $\checkmark$                       & $\checkmark$      & 15.4                  & \textbf{31.2}         & 10.2                  & 20.3                 & 30.1                 & 15.3                 \\
F   & G    & $\checkmark$                      & $\checkmark$ & $\checkmark$                       & $\checkmark$      & 14.0                  & 31.0                  & 9.0                   & 20.9                 & 30.5                 & 15.8                 \\
F   & CG   & -                        & $\checkmark$ & $\checkmark$                       & $\checkmark$      & 14.5                  & 31.0                  & 9.4                   & 22.0                 & \textbf{31.0}        & 17.0                 \\
S   & CG   & $\checkmark$                      & $\checkmark$ & $\checkmark$                       & $\checkmark$      & 5.8                   & 27.1                  & 3.2                   & 13.5                 & 27.6                 & 8.9                  \\
F   & CG   & $\checkmark$                      & $\checkmark$ & $\checkmark$                       & $\checkmark$      & \textbf{15.8}         & \textbf{31.2}         & \textbf{10.5}         & \textbf{22.6}        & \textbf{31.0}        & \textbf{17.7}        \\ \midrule
\multicolumn{6}{c|}{Second Stage}                                                  & hPQ                   & sPQ                   & uPQ                   & hPQ                  & sPQ                  & uPQ                  \\ \midrule
F   & CG   & $\checkmark$                      & -   & $\checkmark$                       & $\checkmark$      & \multirow{3}{*}{15.8} & \multirow{3}{*}{31.2} & \multirow{3}{*}{10.5} & 21.9                 & 31.1                 & 16.9                 \\
F   & CG   & $\checkmark$                      & \multicolumn{2}{c}{PADing}      & $\checkmark$      &                       &                       &                       & 20.6                 & 30.4                 & 15.6                 \\
F   & CG   & $\checkmark$                      & $\checkmark$ & -                         & $\checkmark$      &                       &                       &                       & 20.8                 & 30.6                 & 15.8                 \\ 
\midrule
\multicolumn{6}{c|}{Third Stage}                                                   & hPQ                   & sPQ                   & uPQ                   & hPQ                  & sPQ                  & uPQ                  \\ \midrule
F   & CG   & $\checkmark$                      & $\checkmark$ & $\checkmark$                       & -        & 15.8                  & 31.2                  & 10.5                  & 9.8                  & 11.4                 & 8.7                  \\ \bottomrule
\end{tabular}
}
\label{ablation study}
\vspace{-15pt}
\end{table}

\subsection{Ablation Study}
Note that the experiments employ MaskFormer \cite{maskformer} as the feature extractor trained 12 epochs in all stages and only on panoptic segmentation as its difficulty. The ablation studies on our methods under both inductive and transductive settings are shown in Tab. \ref{ablation study}.  In the inductive setting, only the first stage is processed, and access to both $\textbf{A}_u$ and images with $\textbf{A}_u$ is restricted. As observed in the table, removing CON results in 10.9\% in hPQ, sPQ achieving 31.0\%, and uPQ reaching 6.6\%. Aligning semantic queries with the CLIP visual CLS token without any condition ($\gamma = 0$) significantly boosts hPQ to 15.4\%. However, introducing conditions ($\gamma = 1$) slightly decreases performance to 14.5\% due to a decrease in uPQ. The inclusion of CIA leads to a subsequent increase in hPQ to 15.8\%, reaching its peak. Note that if CGA is replaced with GA, the performance of hPQ even drops to 14.0\%, and if CIA is removed, the hPQ also drops to 14.5\%. If we replace the focal loss with cross-entropy, the hPQ drops drastically to 5.8\% due to the overfitting on seen categories, \ie, 27.1\% sPQ and only 3.2\% uPQ.

\begin{table}[t]
\setlength{\tabcolsep}{20pt}
\caption{Ablation study on the size of the token bank.}
\vspace{-10pt}
\resizebox{\linewidth}{!}{
\begin{tabular}{c|ccc|ccc}
\toprule
\multirow{2}{*}{Bank size} & \multicolumn{3}{c|}{Inductive}                & \multicolumn{3}{c}{Transductive}              \\ \cmidrule{2-7} 
                           & hPQ           & sPQ           & uPQ           & hPQ           & sPQ           & uPQ           \\ \midrule
0                         & 15.3 & \textbf{31.6}          & 10.1 & 22.5          & 30.9          & \textbf{17.7}         \\
16                         & \textbf{16.2} & 31.1          & \textbf{10.9} & 19.8          & 30.4          & 14.7          \\
32                         & 15.8          & 31.2 & 10.5          & \textbf{22.6} & \textbf{31.0} & \textbf{17.7} \\
64                         & 14.6          & 30.9          & 9.6           & 20.7          & 30.4          & 15.7          \\ \bottomrule
\end{tabular}
\label{token bank size}

}
\vspace{-10pt}
\end{table}

\begin{figure}[tb]
  \centering
  \begin{subfigure}{0.35\textwidth}
    \includegraphics[width=\linewidth]{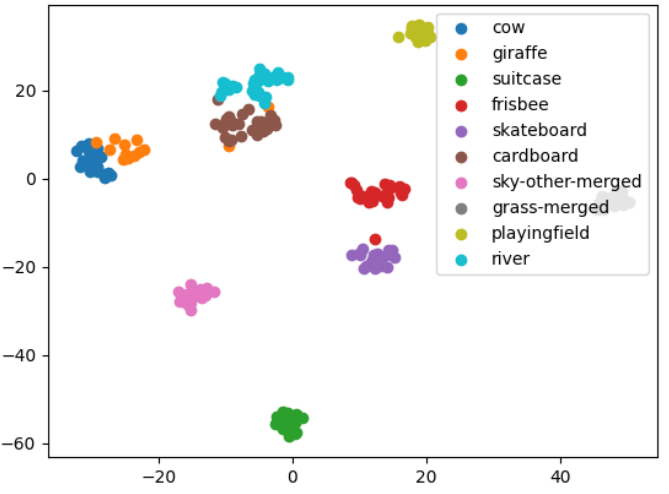}
    \caption{PADing \cite{pading} in the 2nd stage.}
    \label{pading generation}
  \end{subfigure}
  \hspace{30pt}
  \begin{subfigure}{0.35\textwidth}
    \includegraphics[width=\linewidth]{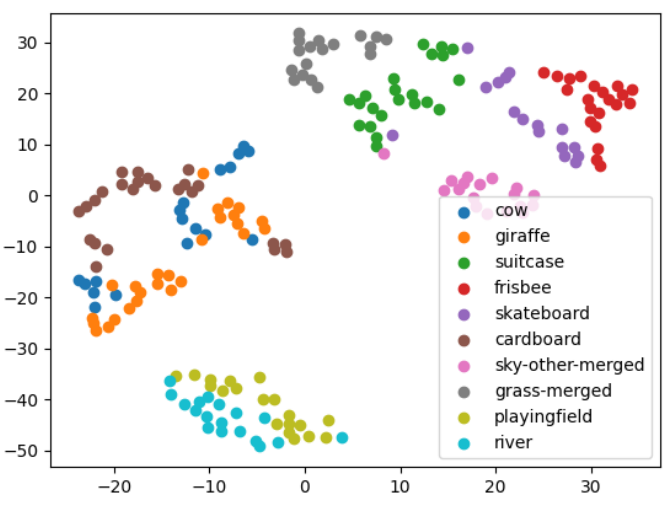}
    \caption{Ours without vision-semantic.}
    \label{nocls generation}
  \end{subfigure}
  \medskip
  \begin{subfigure}{0.35\textwidth}
    \includegraphics[width=\linewidth]{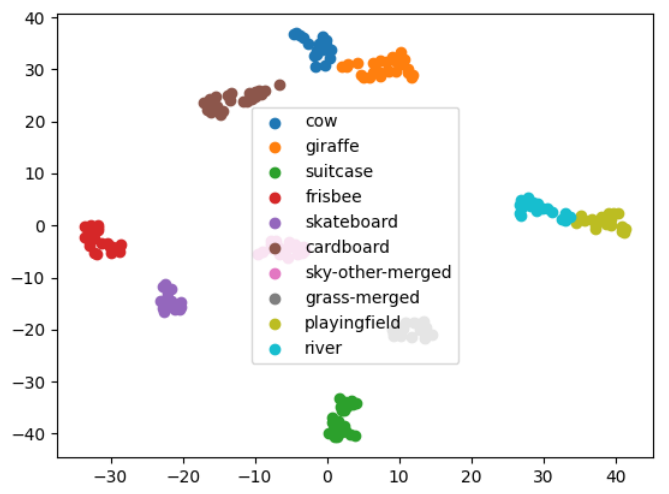}
    \caption{Ours without query contrast.}
    \label{nosimi generation}
  \end{subfigure}
  \hspace{30pt}
  \begin{subfigure}{0.35\textwidth}
    \includegraphics[width=\linewidth]{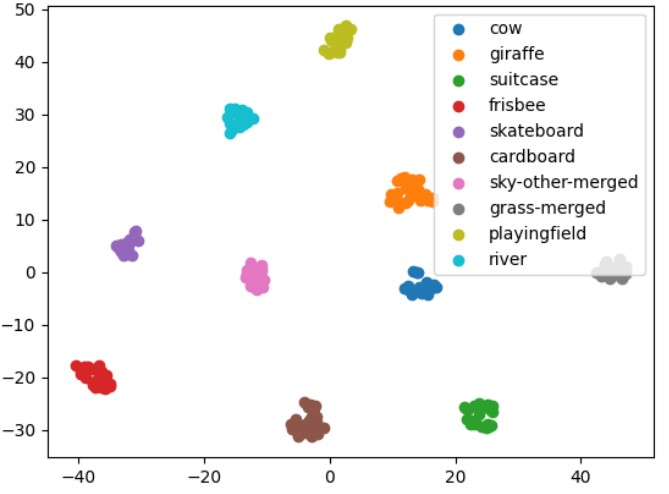}
    \caption{Our generation}
    \label{our generation}
  \end{subfigure}
  \vspace{-10pt}
  \caption{Distribution of the generated pseudo semantic queries.}
  \label{all pseudo generation}
\vspace{-15pt}
\end{figure}

Note that for transductive settings, where all three stages are executed, as different parameters leads to different convergence speed, we only report the best performance of each hyperparameter. If only focal loss is applied, the hPQ reaches 18.1\% with a notable increase in uPQ from 6.6\% to 13.2\%. However, sPQ experiences a significant drop from 31.0\% to 28.7\%. Add the global alignments results in hPQ reaching 20.3\%, with a minimal decrease in sPQ (from 31.2\% to 30.1\%) and a 2.1\% increase in uPQ compared to the baseline method. When we add the CIA, hPQ increases from 20.3\% to 20.9\%, while maintaining sPQ and increasing uPQ to 15.8\%. Introducing conditions in this transductive setting boosts hPQ significantly from 20.3\% to 22.0\%, while maintaining sPQ and increasing uPQ to 17.1\%. Incorporating all proposed methods further enhances hPQ to 22.6\% without compromising sPQ, and uPQ improves to 17.7\%. If softmax is applied, the final hPQ achieves only 13.5\%, indicating our effectiveness.


\noindent \textbf{The size of the token bank.} We further conduct experiments to investigate the effectiveness of the size of the token bank as shown in Tab. \ref{token bank size}. For the inductive settings, we set the size of the bank to 16, 32, and 64. The hPQs achieve 16.2\%, 15.8\%, and 14.6\%, respectively. The sPQ remains relatively stable at 31.1\%, 31.2\%, and 30.9\%. However, for uPQ, with bank sizes of 16 and 32, the uPQs are 10.9\% and 10.5\%, respectively. If the bank size is 64, the uPQ drops significantly to 9.6\%. In the transductive settings, with a bank size of 32, hPQ reaches its peak at 22.6\%, accompanied by a slight decrease in sPQ (from 31.2\% to 31.0\%) and a substantial improvement in uPQ (from 10.5\% to 17.7\%). With a bank size of 64, hPQ continues to increase (from 14.6\% to 20.7\%), but the absolute hPQ remains lower than that of 32. With a bank size of 16, hPQ drops to 19.8\% due to a decrease in uPQ. This experiment underscores the significant impact of bank size on performance. When the bank size is set as 0, the hPQ in inductive reaches 15.3\% due to the decrease of uPQ. However, the performance for transductive is good, \ie, hPQ reaches 22.5\%. To achieve the balance between the inductive and transductive settings, the bank size needs careful design.

\begin{table}[t]
\caption{Comparison on open-vocabulary semantic segmentation task.}
\vspace{-10pt}
\setlength{\tabcolsep}{10pt}
\resizebox{\linewidth}{!}{
\begin{tabular}{ccc|ccc|c}
\toprule
Method    & BackBone                     & Training Dataset                & VOC-20 & PC-59 & A-150 & FPS  \\ \midrule
ZS3 \cite{zs3}       & \multirow{4}{*}{ResNet-101 \cite{resnet}}  & PASCAL VOC \cite{voc}                    & 38.3   & 19.4  & -     & -    \\
LSeg \cite{Lseg}     &                              & PASCAL VOC \cite{voc}                      & 47.4   & -     & -     & -    \\
OpenSeg \cite{Openseg}  &                              & COCO \cite{coco}                           & 60.0   & 36.9  & 15.3  & -    \\
OpenSeg \cite{Openseg}  &                              & COCO \cite{coco} + Loc. Narr. \cite{narr}               & 63.8   & 40.1  & 17.5  & -    \\ \cmidrule{1-3}
Zzseg-seg \cite{simplebaseline}& \multirow{3}{*}{ResNet-101c \cite{deeplabev3}} & \multirow{3}{*}{COCO-Stuff-156 \cite{coco}} & 88.4   & 47.7  & 20.5  & 1.11 \\
DeOP \cite{DeOP}     &                              &                                 & 91.7   & 48.8  & \textbf{22.9}  & 4.37 \\
Ours      &                              &                                 & \textbf{93.7}   & \textbf{49.7}  & 20.9  & \textbf{9.24} \\ \bottomrule
\end{tabular}
}
\label{comparison OV}
\vspace{-15pt}
\end{table}

\noindent \textbf{Visualization of the generated semantic queries.} To analyze the generated pseudo semantic queries, we use t-SNE \cite{tsne} to visualize their distribution, as shown in Fig. \ref{all pseudo generation}. Randomly selecting 10 out of 14 unseen categories from the COCO Panoptic dataset, we compare with the PADing \cite{pading} baseline by replacing its second stage with our primitive generator. The results are depicted in Fig. \ref{pading generation}. While our method effectively separates unseen categories, some semantically similar categories like cow and giraffe remain close. Ablating query contrast in CAT makes it hard to distinguish similar categories, and removing the vision-semantic link makes all categories indistinguishable, highlighting our efficiency.

\subsection{Comparison on Open-Vocabulary Segmentation}
Recently, open-vocabulary semantic segmentation \cite{simplebaseline,DeOP} has attracted much attention from researchers due to its generalization to categories that have never been trained. Note that we use the mIoU of all the categories in this experiment as the metric and under the same setting as other methods. We first train our methods on all the images in the COCO-Stuff dataset with only the annotations from seen categories. Then we directly infer on other datasets without fine-tuning, and the results are shown in Tab. \ref{comparison OV}. Compared with the SOTA methods, \ie, Zzseg-seg, our methods achieve higher performance, \ie, 90.8\%, 47.7\%, and 20.7\%, in all the three datasets, \ie, PASCAL VOC \cite{voc}, Pascal Context, and ADE20k \cite{ade20k}. Meanwhile, without relying on CLIP, our method achieves over \textbf{2 times faster} than the DeOP. Experiments show the generability of our methods.

\begin{figure}[t!]
    \centering
    \includegraphics[width=0.8\linewidth]{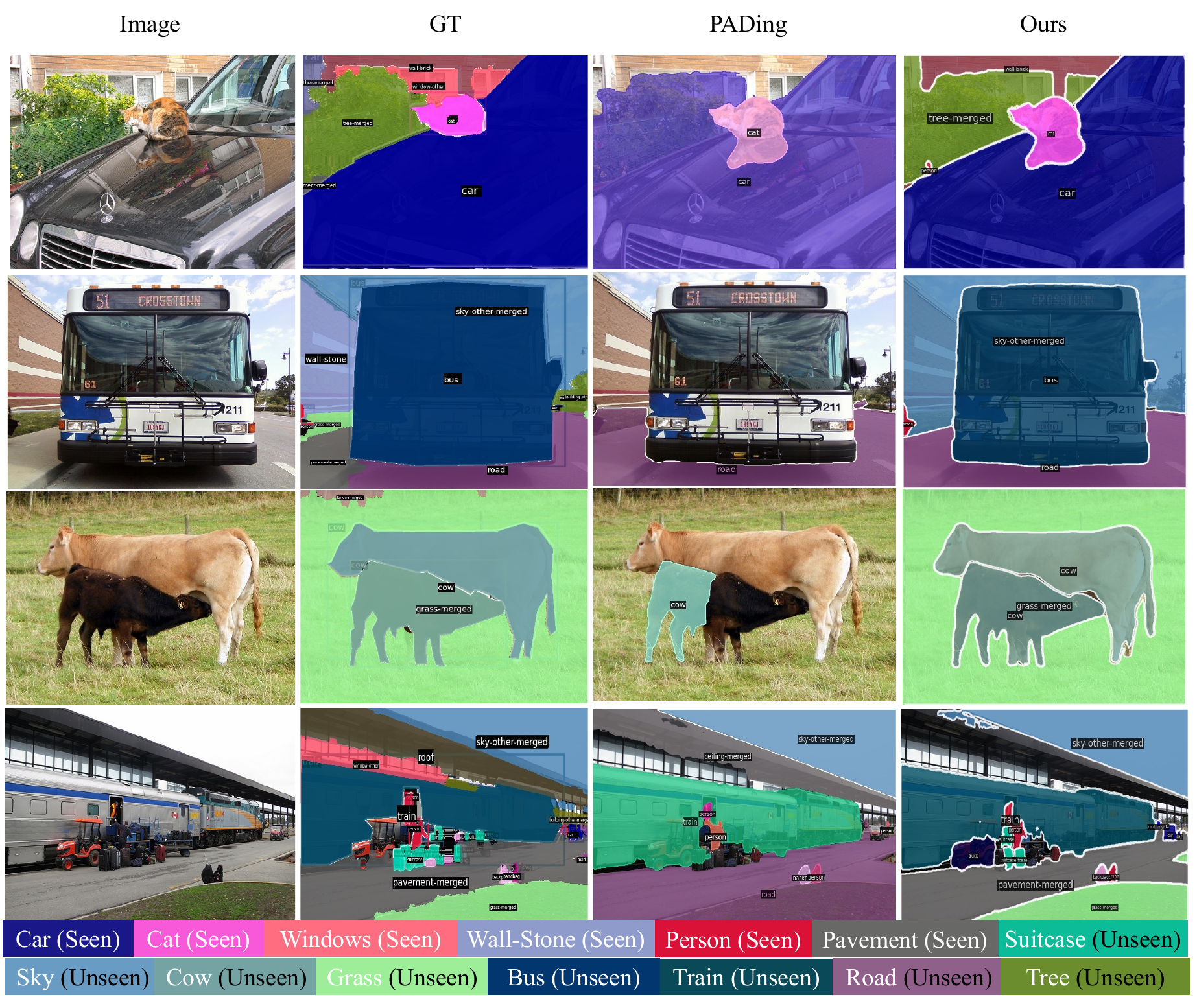}
    \vspace{-5pt}
    \caption{Visualization of our proposed methods.}
    \label{vis figure}
\vspace{-20pt}
\end{figure}

\subsection{Visualization.} 
To qualitatively express the merits of the proposed methods, we visualize the predictions of our proposed methods as shown in Fig. \ref{vis figure}. Our method can achieve better results, especially on the unseen categories.

\section{Conclusion}
We propose \textbf{C}onditional t\textbf{O}ken \textbf{A}lignment and \textbf{C}ycle tr\textbf{A}nsis\textbf{T}ion, \ie, CONCAT, a novel two-stage method that combines the merits of existing projection and generation-based to handle the zero-shot panoptic segmentation. CON provides a shared space with rich semantics and CAT helps generate pseudo unseen queries with high quality. Specifically, CON aligns the semantic queries and visual CLS tokens extracted from complete and masked images by CGA and CIA respectively. Secondly, to generate pseudo queries with high quality, we propose CAT. Formally, CAT is split into semantic-vision and vision-semantic. In semantic-vision, we propose query contrast to model the high granularity of vision features. In vision-semantic, the generated pseudo queries are mapped back to the semantic space and supervised by real semantic embeddings. Finally, we finetune the semantic projector to fit the unseen categories. Compared with SOTA methods, ours achieves higher performance and shows great generability.

\noindent \textbf{Limitations and border impacts.} While our method shows impressive performance, it still can be improved in handling unseen categories, particularly in comparison to seen ones. Moreover, our approaches exhibit sub-optimal performance in open-vocabulary tasks when dealing with large numbers of categories.

\clearpage  

%
%
\bibliographystyle{splncs04}
\bibliography{egbib}
\end{document}